\documentclass[
]{ceurart}

\usepackage{listings}
\usepackage{subcaption}
\begin{document}

\copyrightyear{2023}
\copyrightclause{Copyright for this paper by its authors.
  Use permitted under Creative Commons License Attribution 4.0
  International (CC BY 4.0).}

\conference{CHR 2023: Computational Humanities Research Conference, December 6
  -- 8, 2023, Paris, France}

\title{Blind Dates: Examining the Expression of Temporality in Historical Photographs}

\author{Alexandra Barancová}[
email=a.barancova@uva.nl,
]
\fnmark[1]
\author{Melvin Wevers}[
email=m.j.h.f.wevers@uva.nl,
orcid=0000-0001-8177-4582,
url=http://www.melvinwevers.nl,
]
\fnmark[1]
\cormark[1]
\author{Nanne {van Noord}}[
email=n.j.e.vannoord@uva.nl,
orcid=0000-0002-5145-3603,
url=https://nanne.github.io/,
]
\fnmark[1]
\address{University of Amsterdam}

\cortext[1]{Corresponding author.}
\fntext[1]{These authors contributed equally.}

\begin{abstract}
  This paper explores the capacity of computer vision models to discern temporal information in visual content, focusing specifically on historical photographs. We investigate the dating of images using OpenCLIP, an open-source implementation of CLIP, a multi-modal language and vision model. Our experiment consists of three steps: zero-shot classification, fine-tuning, and analysis of visual content. We use the \textit{De Boer Scene Detection} dataset, containing 39,866 gray-scale historical press photographs from 1950 to 1999. The results show that zero-shot classification is relatively ineffective for image dating, with a bias towards predicting dates in the past. Fine-tuning OpenCLIP with a logistic classifier improves performance and eliminates the bias. Additionally, our analysis reveals that images featuring buses, cars, cats, dogs, and people are more accurately dated, suggesting the presence of temporal markers. The study highlights the potential of machine learning models like OpenCLIP in dating images and emphasizes the importance of fine-tuning for accurate temporal analysis. Future research should explore the application of these findings to color photographs and diverse datasets.
\end{abstract}

\begin{keywords}
  Image dating \sep
  Computer vision \sep
  Temporal analysis \sep
  Historical photographs \sep
  OpenCLIP
\end{keywords}

\maketitle
\section{Introduction}

Time plays a crucial role in shaping our understanding and interpretation of the world around us. Our perception of duration, the sequence of our memories, and the authority with which historical records organize the past all contribute to our lived experiences and memory. This perception extends to our interpretation of visual content, where an image's materiality, content, and style can convey critical temporal information. Despite its significance, this aspect of image understanding remains underexplored in artificial intelligence research~\cite{palermo_dating_2012,van_noord_analytics_2022}.

AI models, typically trained on limited data periods, possess a narrow understanding of temporality due to their lack of `awareness' of historical variations. Although efforts have been made to integrate historical data into language models~\cite{manjavacas2021macberth} and even encode time explicitly~\cite{rosin2022time}, these methods primarily focus on text, leaving visual content interpretation largely uncharted territory.
In this paper, we experiment with the task of `dating' images, predicting when an image was taken based on its visual content.\footnote{All code used for this study is available on GitHub: \url{https://github.com/CANAL-amsterdam/dating-images/}} We examine how different image aspects influence a multimodal AI model's predictive accuracy, uncover structural biases in pre-trained computer vision models, and explore their effects on predictions. Our research aims to extend our understanding of the visual representation of time and its influence on image interpretation. This experiment is situated within a broader goal of developing more temporally-aware computer vision and multimodal models. For this, cross-pollination between AI and humanities scholarship on cultural heritage, archiving, and temporality will be needed; as~\cite{fiorucci_machine_2020} show, interdisciplinary work in this area has been limited, yet it has the potential to be mutually beneficial.

\section{Background}

The challenge of automating dating has been addressed across a variety of historical objects, spanning from photographs~\cite{palermo_dating_2012} and artworks \cite{Mensink_Van_Gemert_2014, Khan_van_Noord_2021} to archaeological sites~\cite{klassen_semi-supervised_2018, toner2019language}. With the increasing digitization of historical documents, many of which lack publication dates, computational methods have been employed to estimate their creation dates, primarily analyzing writing styles~\cite{hamid_deep_2019}, focusing on both the text and the visual content of the writing. 
The automatic dating of historical photographs offers substantial value to archives and museums, but also domains like temporal forensic analysis, where dating can serve as evidence. Forensic applications typically concentrate on an image's material aspects, using techniques that identify specific camera models or devices~\cite{mao_device_2009, ahmed_temporal_2020}. While such methods may be overly meticulous for large-scale dating of historical images, they underscore the importance of material information in establishing an image's capture date.

Beyond material aspects, others show that low-level image features like RGB color derivatives and color angles carry temporal information. Models trained on these features often surpass human accuracy in dating photographs~\cite{palermo_dating_2012, fernando_color_2014}. 
Research in this domain has seen the adoption of neural networks for dating photographs, treating it as an ordinal~\cite{liu_deep_2017}, regression~\cite{jose_when_2017}, classification~\cite{salem_analyzing_2016, ginosar_century_2017, sun_understanding_2022}, or retrieval task~\cite{llados_date_2021}.
Studies have also started to pay attention to image content, emphasizing the connection between time and visual elements or semantic cues. Research indicates that temporal cues can be derived from human appearance features, such as clothing, hairstyles, and glasses~\cite{salem_analyzing_2016, ginosar_century_2017}, or even from architectural elements like windows to estimate the age of buildings~\cite{sun_understanding_2022}. A recent study on a family photo album dataset found that the accuracy of the model used for the dating task improved as the number of faces and/or people in a photograph increased~\cite{stacchio_imago_2022} -- this suggests that certain high-level image features and visual elements may carry more temporal information than others. Finally, ~\cite{chen_whats_2023} recently turned to generative approaches to synthesize portrait images for specific decades between 1880 and the present day to distinguish visual markers for these periods. 

As we transition our focus from photograph materiality to content, an essential challenge arises: deciphering how models interpret higher-level input features to predict dates. This exploration aims to yield deeper insights into the ways in which temporality is encoded in visual content and how we can enhance computer vision systems' ability to interpret cultural artifacts.

\section{Image Dating}

We use OpenCLIP~\cite{ilharco_openclip_2021}, the open source implementation of CLIP~\cite{radford_learning_2021}, to predict when a photograph was taken. CLIP is a multi-modal language and vision model that has been shown to have a strong zero-shot capability on diverse vision tasks~\cite{radford_learning_2021} and to outperform a number of domain-specific models on various vision and language tasks following task-specific fine-tuning~\cite{shen_how_2021}. Our interest lies in understanding how visual features, particularly objects, are leveraged for dating purposes, while also evaluating the model's aptitude for the dating task. Among the various models that have been used for dating images, we have not yet seen experiments with large models that have zero-shot capabilities. Experimenting with the potential of such models is interesting due to the lesser need for training data, their broader generalizability and the possibility to examine a multimodal, based on textual and visual data, perspective on tasks like dating.

\begin{figure}[]
\centering
\includegraphics[scale=0.5]{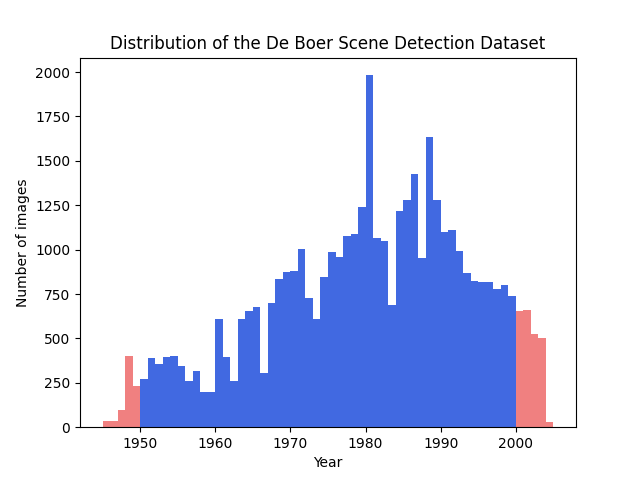}
\caption{Distribution graph of the \textit{De Boer Scene Detection} dataset by year. We utilized the data marked in blue in our experiments.}
\label{fig_dataset_distribution}
\end{figure}

\paragraph{Data} Collections of press photographs are available with relatively reliable dates, making them well-suited for examining the visual representation of time. For this experiment, we have chosen to use the \textit{De Boer Scene Detection} dataset, which contains 52,160 digitized historical press photographs from the \textit{De Boer} newspaper agency spanning from 1945 to 2005.\footnote{The dataset is available at \url{https://zenodo.org/record/7137452}} The images are scanned and cropped photo negatives, the vast majority of which are gray-scale.~\cite{WeversEtAl.2022} The dataset contains relatively mundane photographs, rather than iconic ones, as well as a wide variety of different scenes ranging from sporting events to landscapes~\cite{wevers_scene_2021}; this makes it an interesting case for exploring the visual elements that carry temporal information. Besides the exact year, each image has a label describing the depicted scene. We excluded images lacking date information and those taken before 1950 and after 1999, resulting in 39,866 photographs. These cut-off points were chosen to include data spanning complete decades in our analysis. Figure~\ref{fig_dataset_distribution} illustrates the dataset distribution per year. We split the dataset in a train and test set using stratified sampling based on the year, with the aim of reducing uneven distributions across the splits to prevent biases, in 80\% and 20\% respectively.

We structure our experiment in three steps: examining zero-shot classification capabilities, fine-tuning the model, and assessing the impact of visual content on the model's dating ability. 

\paragraph{Zero-shot Classification.} To investigate to what extent OpenCLIP can be used for dating we apply zero-shot classification to the test set to predict the photograph's date. This process uses the prompt `a photograph from the year $x$' where $x$ ranges from 1950 up to 2000. We employ Mean Absolute Error (MAE) for performance evaluation, following~\cite{fernando_color_2014, jose_when_2017}.
We find an MAE of 15.8, which indicates relatively poor performance considering both the 50 year range of our dataset and the comparison to results others have demonstrated on the dating task ---~\cite{jose_when_2017} and ~\cite{llados_date_2021}, for example, attained MAE values of up to 7.12 and 7.48 respectively in their experiments with the Date Estimation in the Wild (DEW) dataset that covers the period 1930-1999. Additionally, \cite{jose_when_2017} showed that human participants had an MAE of 10.9 on the DEW dataset and as such were on average about 11 years off in their predictions. Comparatively, the almost 16 years that the zero-shot model achieves is quite poor. 

\begin{figure}[]
    \centering
    \begin{subfigure}{0.45\textwidth}
        \centering
        \includegraphics[width=\textwidth]{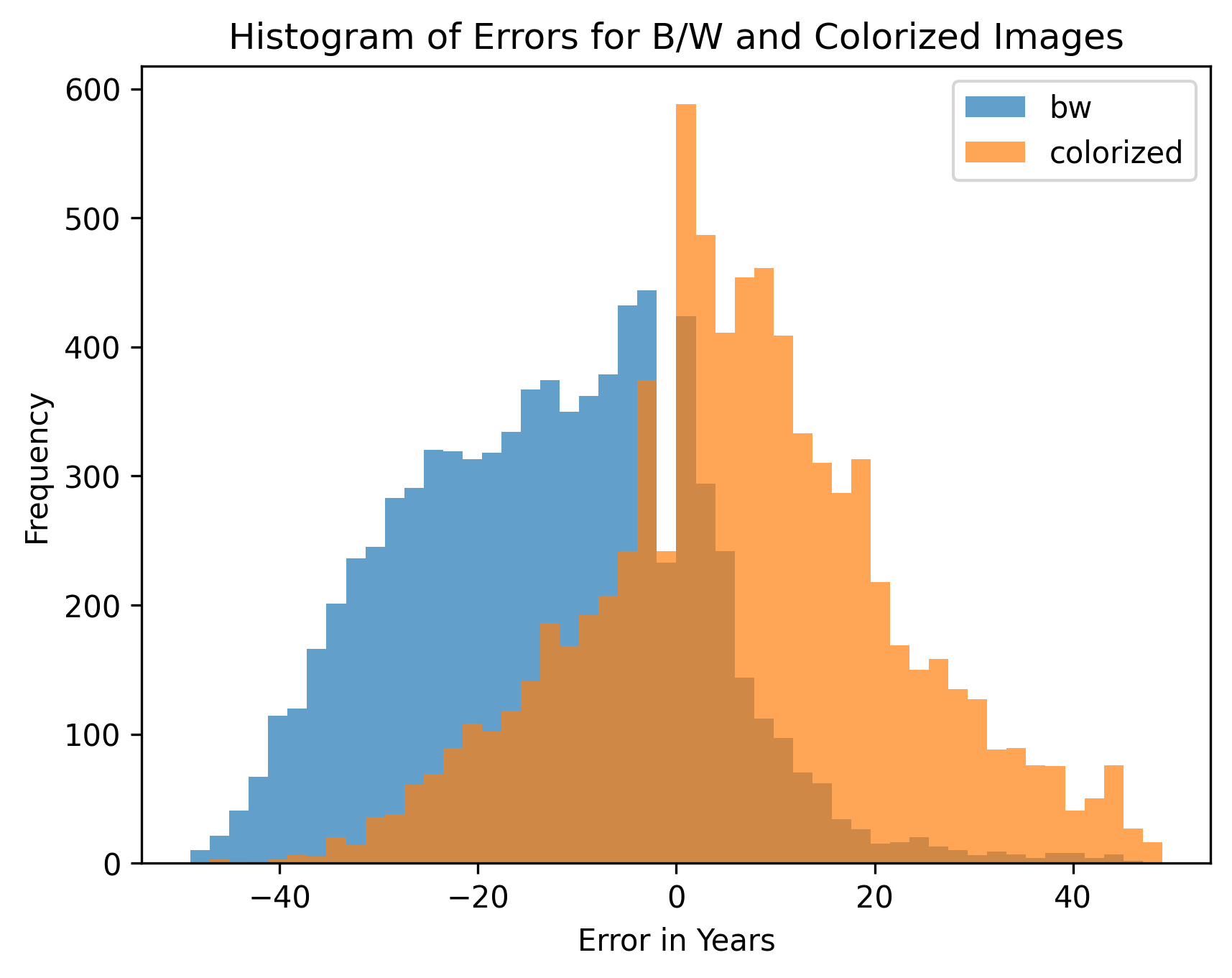}
        \caption{Zero-shot classification}
        \label{fig:sub1}
    \end{subfigure}
    \hfill
    \begin{subfigure}{0.45\textwidth}
        \centering
        \includegraphics[width=\textwidth]{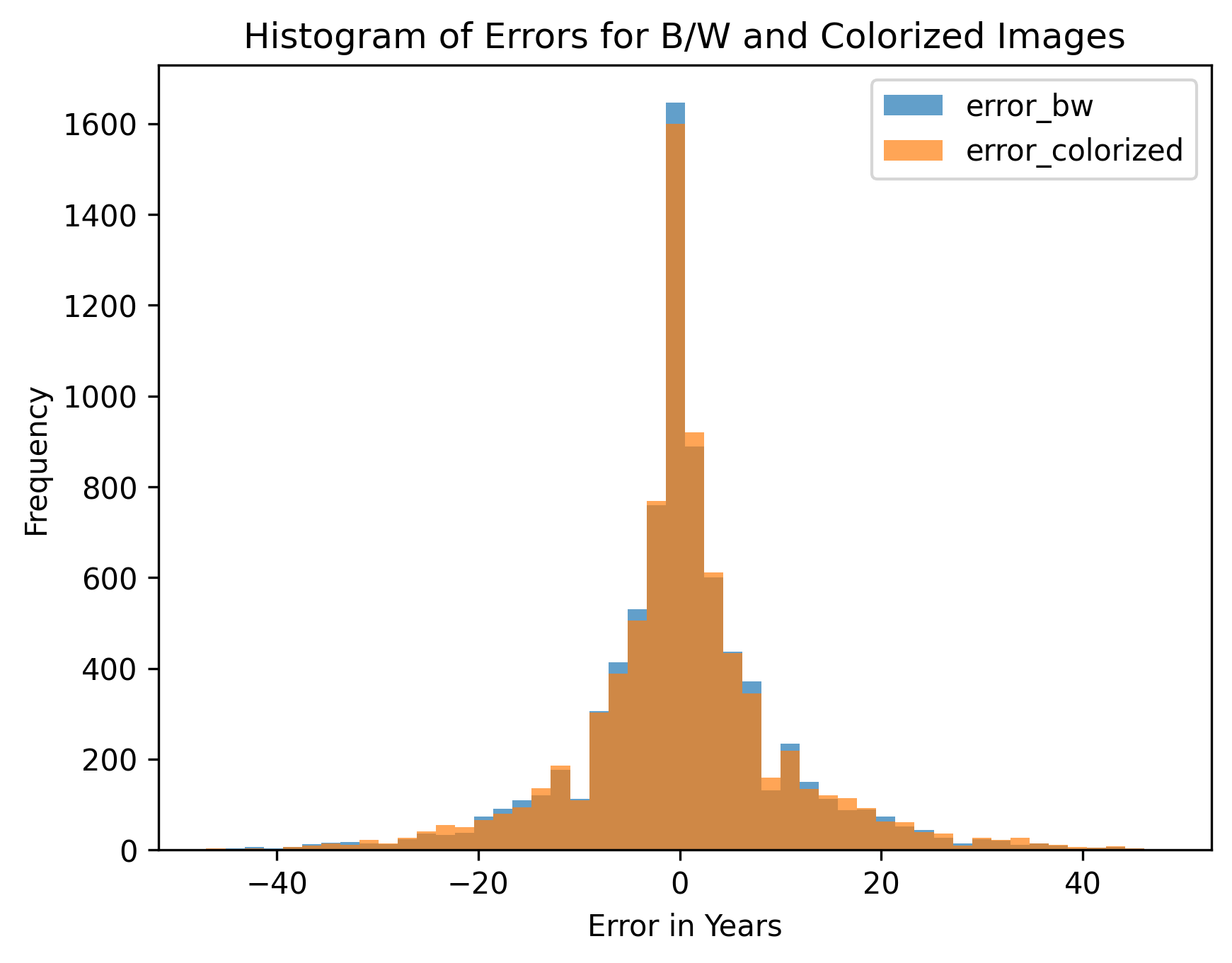}
        \caption{Fine-tuned classification}
        \label{fig:sub2}
    \end{subfigure}
    \caption{Histograms showing the error distribution for the zero-shot and fine-tuned classification.}
    \label{fig:complete}
\end{figure}

\begin{figure}[]
\centering
\includegraphics[width=\textwidth]{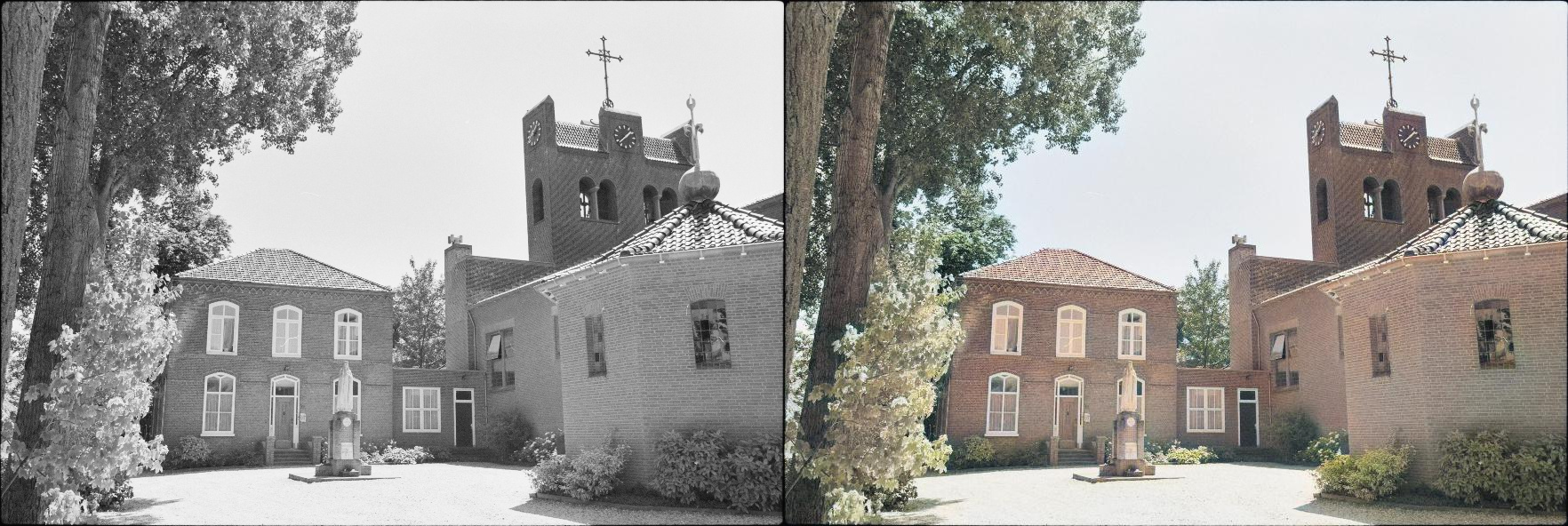}
\includegraphics[width=\textwidth]{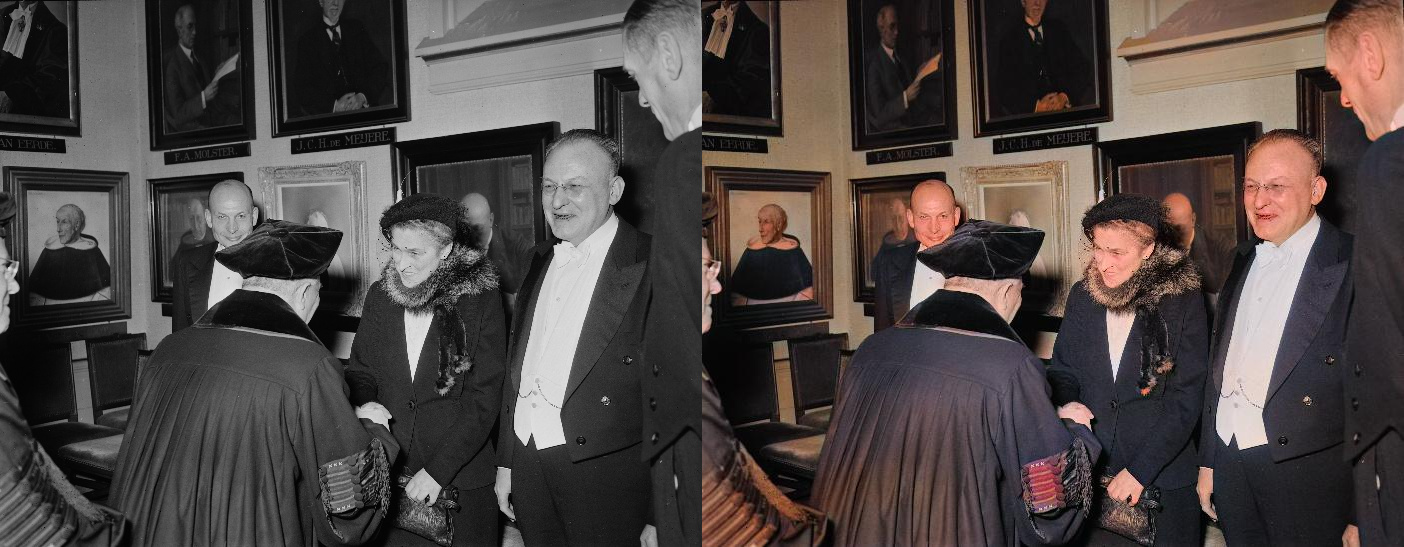}
\caption{Examples of original (left) and colorized (right) photographs from the \textit{De Boer Scene Detection} dataset for which colorization had the largest impact on decreasing error for zero-shot classification. Colorization decreased the error from -37 to 0 for the top image (actual year: 1999, prediction original: 1952, prediction colorized: 1999), and from 37 to 2 for the bottom image (actual year: 1952, prediction original: 1999, prediction colorized: 1954).}
\label{fig:sample_images}
\end{figure}

The error rate distribution reveals a preference in the zero-shot model for predicting dates earlier than the actual date. Suspecting a correlation with the images' gray-scale nature, we tested the zero-shot classification with a colorized version of our dataset,\footnote{We colorized all the photographs using DeOldify \url{https://github.com/jantic/DeOldify}} resulting in the error distribution leaning slightly more toward the future (see Figure~\ref{fig:sub1}).\footnote{A KS-test (KS-statistic: 0.49, p-value: 0.0) supports the difference between the distributions.}
Figure~\ref{fig:sample_images} shows two sample images for which colorization had a large effect on decreasing date prediction error. One depicts an outdoor view of a church, the other a group of people in formal wear. Both images show a large prediction error in the gray-scale variant, albeit in different directions, i.e. overestimate or underestimating the actual date. 
We find that colorizing the images improves the overall zero-shot capabilities of OpenCLIP, however, with an MAE of 13.2 it is still relatively ineffective for the dating task.

\paragraph{Fine-tuned Classifier.}

To overcome the zero-shot limitations we explore whether fine-tuning OpenCLIP improves performance on the dating task and eliminates the bias found in the initial experiment. To this end, we train a logistic classifier using the OpenCLIP image embeddings. Training a logistic classifier also allows us to focus solely on the temporal information in the visual content, removing possible confounding temporal bias in the model from text, through prompting. When using textual prompts for zero-shot classification, the words used might be better suited for specific historical periods, thereby introducing a bias in the images corresponding to this prompt. 
Fine-tuning reduced the error and the bias, with the MAE being 6.65 for the classifier trained and evaluated on the original gray-scale images and 6.79 for the colorized images. The bias between gray-scale and colorized images found in the zero-shot approach disappears after fine-tuning (Figure~\ref{fig:complete}), displaying a more normal error distribution.\footnote{Also supported by KS-test (K-S statistic: 0.00, p-value: 0.92).}

\paragraph{Content Analysis.}
Upon training a model to predict dates, we investigated the content of the images to determine whether specific visual features improved or hindered the predictions. An initial analysis using the available scene labels proved to be inconclusive as there were large differences in the within-scene error rates, as well as a large variety of visual features represented in individual scenes. We opted to examine the visual features at the object level, using Detectron2 outputs~\cite{wu2019detectron2}. 
For the detection task, we used the 80 default objects, as defined in COCO, default ROI threshold (0.5). Next, we only selected detection with a confidence threshold above 0.8 and types that appeared more than 200 times in the entire data set, in order to shorten the long tail. The confidence threshold was output by Detectron2 per image. Finally, we picked 12 classes representing modes of transport and living beings to focus this experiment on.\footnote{`bicycle', `boat', `bus', `car', `motorcycle', `train', `truck', `bird', `cat', `dog', `horse', `person'}. Our motivation in picking these classes was to reduce the granularity of the available categories so as to identify larger trends; classes like `tie` for example, might be closely related to `person', essentially functioning as a sub-class thereof. An additional motivation for excluding some of the MS COCO object classes, is that they did not all suit the context and/or time span of our dataset, especially technology like `laptop', `cell phone' or `microwave'.

A Bayesian regression analysis was conducted to measure the effect of object presence and absence on the error rate.\footnote{The analysis was performed using the Python library Bambi and the NumPyro nuts sampler.} The regression model was defined as follows:

\[ prediction\_error = 1 + object\_presence\]

Where $prediction\_error$ is the outcome variable, $object\_presence$ is the binary prediction variable indicating each object's presence. To model the distribution, we assumed a negative binomial. We model the errors as counts, where the event is the counts of predictions with a specific error.\footnote{Since the variance and mean are not equal a zero-inflated Poisson was not warranted. See the GitHub for more information on the models.}

\begin{figure}[htbp]
  \centering
  \begin{subfigure}[t]{0.45\textwidth}
    \centering
    \includegraphics[width=\linewidth]{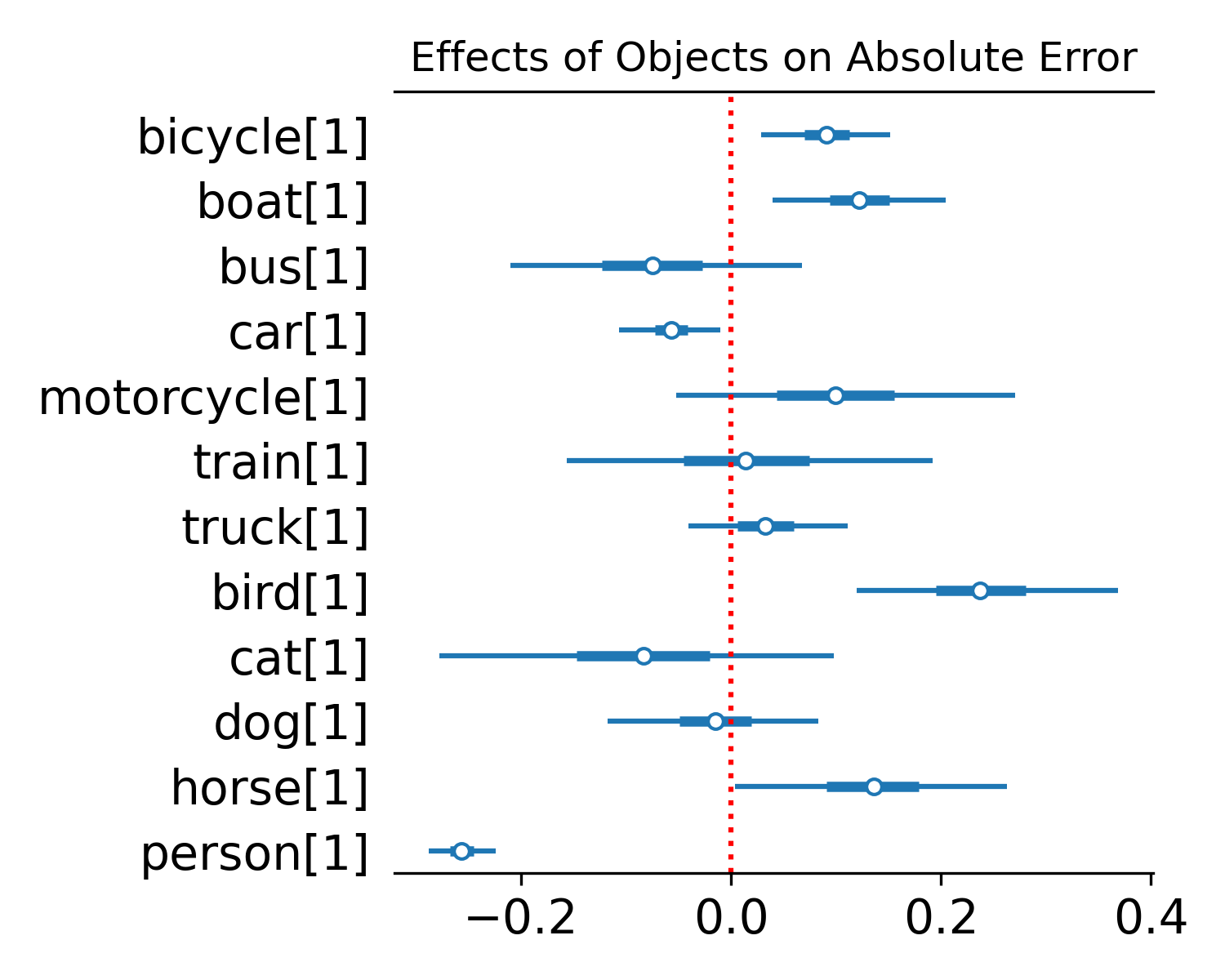}
    \caption{Estimated effects of objects on absolute error. HDI .95, meaning that there is a 95\% chance that the true value lies within this range.}
    \label{fig:resultsa}
  \end{subfigure}
  \hfill
  \begin{subfigure}[t]{0.45\textwidth}
    \centering
    \includegraphics[width=0.8\linewidth]{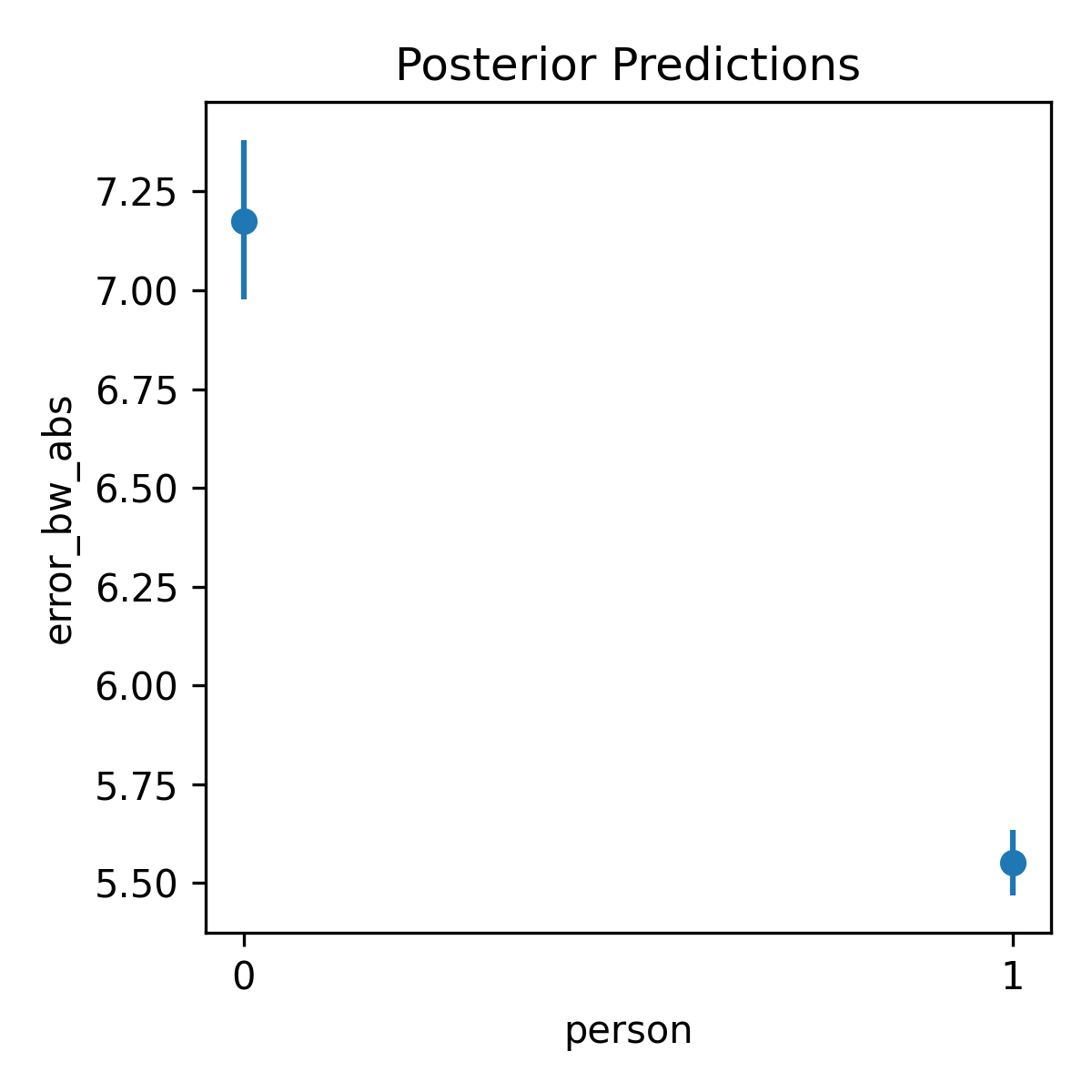}
    \caption{Posterior predictions for the class `person'. A person in the image reduces the absolute error from 7.2 to 5.5.}
    \label{fig:resultsb}
  \end{subfigure}
  \caption{Output of the regression model. The MAE values are based on the entire dataset, without taking into account random effects, which results in higher reported MAE values. However, for this analysis the change in MAE is of primary interest. }
  \label{fig:results}
\end{figure}

Figure~\ref{fig:resultsa} shows that for modes of transport, the presence of `bicycle', `boat', `motorcycle', and `train' increase the absolute error, whereas `bus' and `car' decrease the absolute error. We hypothesize that these vehicles are more prone to exterior changes in this period. Of the animals, we see that `bird' and `horse' increase the error and `cat' decreases the error. Our hypothesis here is that cats might be depicted more often together with humans and in interior environments, which may include more temporal markers. Finally, we see that having a `person' in an image has the strongest effect, decreasing the MAE from approximately 7.2 to 5.5 (Figure~\ref{fig:resultsb}), indicating that depictions of people convey visual cues about time. These results and hypotheses need further examination, which we intend to undertake in future work.

\section{Conclusion and Discussion}

Our exploration with OpenCLIP to date historical press photographs yielded several findings. 

\paragraph{Ineffectiveness of Zero-shot Classification.} Our first finding is that the zero-shot classification capability of OpenCLIP does not perform well in dating images. The model demonstrated a distinct bias towards predicting earlier dates, which we attribute to the gray-scale nature of our images. This suggests that OpenCLIP may have learned to associate gray-scale with older photographs, as~\cite{offert_concept_2023} also concluded in exploring the concept of history in foundation models including CLIP. We attempted to counteract this bias by colorizing the images, which mildly improved the model's accuracy and shifted the bias towards predicting more recent dates. However, despite these adjustments, the efficacy of zero-shot classification for this task remained limited.

\paragraph{Improvement through Fine-tuning.}
Our second finding is that fine-tuning OpenCLIP using a logistic classifier significantly enhances the model's performance. The fine-tuned model effectively eliminates the bias towards past dates seen in the zero-shot approach and offers comparable accuracy levels for both gray-scale and colorized images. This indicates that the presence of color in images becomes less significant for dating them when the model is trained to focus on visual content. Future research could look into generating captions rather than labels or scenes to provide a more enriched context for each image.

\paragraph{Objects as Temporal Markers.} 
Our third finding, coming from the post-hoc regression analysis, is that the presence of people in images generally leads to more accurate date predictions. This echoes the findings of~\cite{stacchio_imago_2022}. We posit that this could be attributed to the time-dependent markers humans tend to carry, like fashion and hairstyles, as has previously been shown in studies on yearbook portraits by ~\cite{ginosar_century_2017} and~\cite{salem_analyzing_2016}. Moreover, we see that the presence of animals often kept as house pets also reduces the error, we hypothesize that this might be due them being photographed indoors or in proximity to humans, which might carry more temporal markers than animals, such as horses or birds that are captured in nature. Finally, certain modes of transportation increase the error rate while others decrease it. We need to explore to what extent this is related to innovations that lead to visual changes over time. 
All in, further investigation is necessary to validate these hypotheses; considering our findings on the influential role of human figures in the images, it would be worthwhile to explore datasets containing fewer human figures -- in our dataset only 8,674 of the 39,866 photographs contained no people.\footnote{This is based on outputs from Detectron2 at an object confidence threshold of 0.8.} This could shed light on whether the presence of humans is generally advantageous for image dating, or if this is a specific characteristic of our dataset or a manifestation of model bias. 

\vspace{\baselineskip}

\noindent In conclusion, this study deepens our understanding of how computer vision models interpret and extract temporal information from historical visual material. It highlights the potential of OpenCLIP for image dating tasks. It also underscores the importance of model fine-tuning to counter biases. Future work could test our findings' generalizability to color images and datasets from various periods and geographical regions. Such work can be a means to identifying and using temporal information in visual material better, with the aim of creating more temporally-aware computer vision and multimodal models. To this end, we see case studies engaging with specific computer vision/ temporal tasks like image dating, as important steps in testing what works in terms of both models and data.

\bibliography{sample-ceur}

\end{document}